\documentclass[letterpaper, 10 pt, conference]{ieeeconf}  

\IEEEoverridecommandlockouts                              

\usepackage{amsmath} 
\usepackage{amssymb}  
\usepackage{graphicx}
\usepackage{subcaption}
\usepackage[ruled, linesnumbered]{algorithm2e}
\usepackage{supertabular, booktabs, multirow, multicol, makecell}
\usepackage{color}
\usepackage{balance}
\usepackage{hyperref}

\newenvironment{small_ind_s_itemize2}{\begin{list}{$\bullet$}
{\setlength{\rightmargin}{0em}
\setlength{\leftmargin}{1em}
\setlength{\itemsep}{0em}
\setlength{\itemindent}{-0.6em}
\setlength{\topsep}{0em}
\setlength{\parsep}{0em}}}{\end{list}}

\newenvironment{small_ind_s_itemize}{\begin{list}{$\bullet$}
{\setlength{\rightmargin}{0em}
\setlength{\leftmargin}{1em}
\setlength{\itemsep}{0em}
\setlength{\topsep}{0em}
\setlength{\parsep}{0em}}}{\end{list}}

\newcounter{ctr}

\newenvironment{small_ind_enumerate}{\begin{list}{\thectr.}
{\usecounter{ctr}
\setlength{\rightmargin}{\rightmargin}
\setlength{\leftmargin}{1.2em}
\setlength{\topsep}{\topsep}
\setlength{\itemindent}{-0.8em}
\setlength{\parsep}{\parsep}}}{\end{list}}

\title{\LARGE \bf
Towards Robust One-shot Task Execution using \\ Knowledge Graph Embeddings
}

\author{Angel Daruna$^{1}$, Lakshmi Nair$^{1}$, Weiyu Liu$^{1}$ and Sonia Chernova$^{1}$
\thanks{$^{1}$Georgia Institute of Technology, Atlanta, GA. Email: {\tt\small \{adaruna3, lnair3, wliu88, chernova\}@gatech.edu}}
\thanks{This work is supported in part by NSF IIS 1564080, NSF GRFP DGE-1650044, and ONR N00014-16-1-2835. Any opinions, findings, and conclusions or recommendations expressed in this material are those of the author(s) and do not necessarily reflect the views of the supporters.}
}

\begin{document}

\maketitle
\thispagestyle{empty}
\pagestyle{empty}

\begin{abstract}

Requiring multiple demonstrations of a task plan presents a burden to end-users of robots. However, robustly executing tasks plans from a single end-user demonstration is an ongoing challenge in robotics. We address the problem of one-shot task execution, in which a robot must generalize a single demonstration or prototypical example of a task plan to a new execution environment. Our approach integrates task plans with domain knowledge to infer task plan constituents for new execution environments. Our experimental evaluations show that our knowledge representation makes more relevant generalizations that result in significantly higher success rates over tested baselines. We validated the approach on a physical platform, which resulted in the successful generalization of initial task plans to 38 of 50 execution environments with errors resulting from autonomous robot operation included.

\end{abstract}

\section{Introduction}
Robust one-shot task execution is an ongoing challenge in robotics. Learning from Demonstration (LfD) provides the means for end-users to program new task plans (structured sequences of abstract primitive actions \cite{ravichandar2020recent}). Typically, multiple varying demonstrations are required in order to learn task plans that are resilient to failures in execution of primitive actions. Requiring multiple demonstrations can present an additional burden to the end-user. However, task plans that are learned in domains rich with semantic knowledge can benefit from incorporating such domain knowledge into the task plan. Task plan constituents from a prototypical or demonstrated task plan can be effectively generalized to new execution environments by leveraging statistical correlations or heuristic features learned from the task domain \cite{boteanu2016leveraging,nyga2018grounding}.

We address the problem of one-shot task execution, in which a robot must generalize a \textit{single} demonstration or prototypical example of a task plan to a new execution environment. For example, the robot could be shown a demonstration of cleaning a kitchen counter using a napkin found in the kitchen cabinet. The robot may have to repeat the task plan in a new environment where napkins may be kept in another location, or are unavailable altogether. Such differences between environments typically halt execution of the task plan due to unsatisfied preconditions when searching for the missing object, causing a primitive action failure.

\begin{figure}[t]
	\centering
	\includegraphics[width=0.49\textwidth]{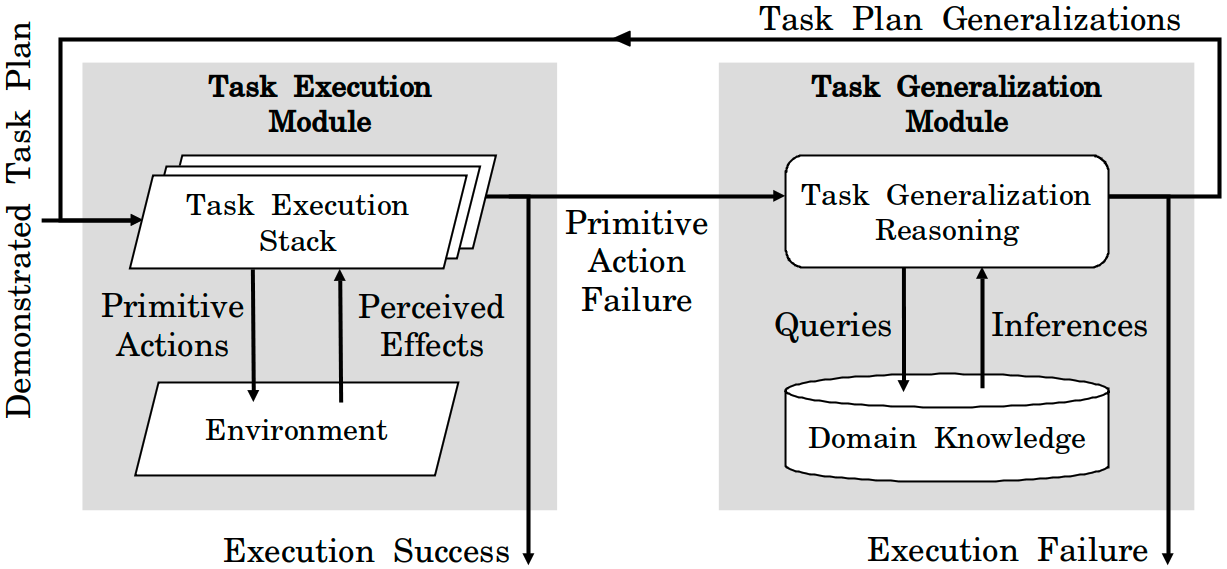}
	\captionsetup{width=\linewidth}
	\caption{\small 
	The task generalization module incrementally generalizes failed task plans by leveraging the learned knowledge graph to infer plan constituents (see Sec~\ref{sec:approach}). The task execution module sequentially executes primitive actions to complete the task plan.}
	\label{fig:Pipeline}
	\vspace{-0.8cm}
\end{figure}

Semantic knowledge tends to have the properties of being \textit{large-scale} and \textit{sparse} (i.e. there are few true facts in a space of many possible facts). We use knowledge graph (multi-relational) embeddings (KGE) as a computational framework to learn and represent task domain knowledge because KGEs are designed for knowledge graphs that are large-scale and sparse \cite{chen2020review}. We integrate KGEs as a knowledge representation along with the task plan, in a \textit{task generalization module}. Our developed task generalization module infers appropriate task plan constituents by reasoning about the learned domain knowledge. Task plan generalizations are then used by a task execution module when the demonstrated task plan fails. The task execution module sequentially executes primitive actions on an environment to complete the task plan. Our implemented architecture (Figure~\ref{fig:Pipeline}), enables one-shot task execution both in simulation and on a physical robot.

We compare our domain knowledge representation against four representative baselines in the context of robust one-shot task execution, including prior KGEs \cite{daruna2019robocse}, Markov Logic Networks \cite{nyga2013pracmln,tenorth2013knowrob}, Plan Networks \cite{orkin2007restaurant}, and word embeddings \cite{fulda2017harvesting}. In simulation, we evaluate each method's ability to generalize an initial demonstrated task plan to 300 execution environments, averaged across 40 different initial demonstrations for a total of 12,000 executions. Our experiments in simulation indicate that our knowledge representation provides improvements in success rate over the baselines with strong statistical significance. We follow these experiments with an ablation study that characterizes the variances of our performance metrics with respect to components of the task generalization module. Lastly, we validate the implemented architecture on a physical robot that must autonomously execute an initial demonstrated task plan in 10 new execution environments, for 5 different initial demonstrations. Our system successfully generalizes the initial task plans to 38 of 50 sampled execution environments, where failures stem solely from robot errors, such as precarious object grasps. Our contributions are summarized as follows.

\begin{small_ind_s_itemize2}
    \smallskip
    \item We design a new robot knowledge graph that adds entity and relation types not present in prior work \cite{daruna2019robocse};
    \item We design a task generalization approach that integrates the KGE computational framework with multiple task reasoning levels to generalize task plans;
    \item We demonstrate our approach's advantages against representative baselines in 12,000 simulations and validate the implemented architecture on a physical robot performing one-shot task execution in 50 execution environments.
\end{small_ind_s_itemize2}

\section{Related Work}
\label{sec:related}

Prior work in LfD has tackled the challenging problem of extracting task plans from a single end-user demonstration \cite{orendt2016robot, bajracharya2019mobile, mohseni2015interactive, rybski2007interactive, friedrich1995robot, balakuntala2019self, goo2019one, duan2017one, yu2018one}. These approaches present intuitive ways for end-users to program complex robot behaviors using kinesthetic teaching \cite{orendt2016robot}, virtual reality \cite{bajracharya2019mobile}, GUI programming \cite{mohseni2015interactive}, or direct demonstration \cite{yu2018one}. However, generalization of task plans in the case of variations between the demonstration and execution environments continues to be a challenging problem. Prior work has focused on task generalization by adapting action trajectories \cite{orendt2016robot}, composing primitive behaviors \cite{yu2018one}, or incorporating decision points into the demonstration \cite{bajracharya2019mobile}. However, these approaches rely primarily on the end-user's demonstrations to achieve robust execution. Instead, our approach integrates domain knowledge with the task plan to avoid relying solely on the end-user's demonstrations. Furthermore, our approach does not assume a fully observed world state, nor that all the objects used during the demonstration are available.

Prior work in open-world planning has developed techniques to make robots programmed by expert users more robust to environment variation through the use of common-sense or domain knowledge \cite{hanheide2017robot, zhang2015mixed, tenorth2009knowrob, saxena2014robobrain, chernova2020situated}. Open-world planners can achieve plan goals in environments with variations \cite{hanheide2017robot}, under-specified plans \cite{tenorth2009knowrob,nyga2017instruction}, and partially observed worlds \cite{zhang2015mixed} by coupling a knowledge base with declarative programming planners, such as PDDL \cite{hanheide2017robot}, CRAM \cite{tenorth2009knowrob}, or ASP \cite{zhang2015mixed}. However, these techniques tend to depend on hand engineered domain knowledge inputs that end-users cannot provide in a new demonstration, such as a taxonomy of relevant concepts or an ontology of the domain.

Most closely related to our approach are \cite{nyga2018grounding} and \cite{boteanu2016leveraging}, which leverage different inputs and computational frameworks to generalize task plans for new execution environments in addition to having differing assumptions. In \cite{nyga2018grounding} online ontologies and repositories of annotated task demonstrations are provided as inputs to learn a probabilistic graphical model, a Markov Logic Network (MLN). The MLN relating objects and object attributes is used to infer the most likely generalization of a prototypical task plan that gets invoked from a natural language command. Generalizations can include alternate primitive actions and primitive action parameters. In \cite{boteanu2016leveraging}, a random forest classifier is trained on similarity features computed over general lexical databases of objects and object attributes provided as an input. The classifier is then used to infer valid object substitutions for objects used in a demonstrated task plan and assumes the same primitive action is being performed. Our approach uses a KGE to learn a vectorized representation of a knowledge graph relating objects and object attributes. Training data provided as input to learn the embedding is mined from simulations without the need for end-users. The learned embedding vectors can then be used to infer likely task plan constituents including primitive actions and primitve action parameters.

\section{Problem Definition}
\label{sec:problem}

Our work is motivated by applications in which a robot operates in dynamic environments, such as households, with naive users who provide a demonstration of a task plan $T_d$. The task plan is defined as a sequence of primitive actions $\{a^1_d, a^2_d, ..., a^k_d\}$, where each primitive action may or may not be parameterized by objects. The execution environment can be different from the demonstration environment in terms of environment state, such as differing object types available for the task. As a result, the demonstrated task plan fails due to unsatisfied pre-conditions of primitive actions, and the robot must generalize the demonstrated task plan to formulate an \textit{executable} task plan $T_x$. We then formulate our problem as:

\smallskip \noindent \textit{\indent Given an execution environment $E_x$ and a task plan $T_d$ recorded from a demonstration in a demonstration environment $E_d$, can a modified task plan $T_x$ be formulated for $E_x$ such that the robot is able to accomplish the task?}

\smallskip 

We assume that unsatisfied pre-conditions stem solely from environment state changes (perturbations) that prevent the completion of a primitive action. Perturbations are limited to varying the type or location of the object available to perform the task. We note that this assumption commonly applies to household settings where objects do not have static locations or objects available for tasks vary between households. We consider other sources of failure, such as hardware failure or manipulation errors, out of scope.

\section{Approach}
\label{sec:approach}

Our approach generalizes a failed demonstrated task plan by reasoning about the primitive actions and the learned domain knowledge, as shown in Figure~\ref{fig:Pipeline}. When the execution of a demonstrated task plan $T_{d}$ is halted due to a failed primitive action, the task generalization module is called to infer a task plan $T_{x}$ for the execution environment $E_{x}$. The task generalization module generalizes the demonstrated task plan by iteratively querying the learned domain knowledge representation and making incremental updates to the task plan. The updated task plan is then sent back to the task execution module to resume the task from the failure point. In particular, we integrate our approach with a task execution module developed in prior work for mobile manipulator robots \cite{banerjee2020taking}. We provide further details on the components of the task generalization module below.
\vspace{-0.1cm}

\subsection{Knowledge Representation}
\label{sec:KGE}
\vspace{-0.1cm}

Our domain knowledge representation uses an explicit model of world semantics in the form of a knowledge graph $\mathcal{G}$ composed of individual facts or triples $(h,r,t)$ with $h$ and $t$ being the head and tail entities (respectively) for which the relation $r$ holds, e.g. {\small$($\textit{cup, hasAction, fill}$)$}~\cite{beetz2018know, chernova2020situated, saxena2014robobrain, zhu2014reasoning}. In this work we model $\mathcal{G}$ using graph embeddings because of their ability to learn the underlying structure of graphs and infer new facts beyond known facts in a graph~\cite{fulda2017harvesting, thomason2018guiding, daruna2019robocse, scalise2019improving, paulius2020motion, arkin2020multimodal}. We build upon the framework in RoboCSE~\cite{daruna2019robocse}, which uses a KGE to represent $\mathcal{G}$. 

KGEs model $\mathcal{G}$ in vector space~\cite{wang2017kge_survey}, learning a continuous vector representation from a dataset of triples $\mathcal{D}\!=\!\big\{(h,r,t)_i,y_i|\,h_i,t_i\!\in\!\mathcal{E},r_i\!\in\!\mathcal{R},y_i\!\in\!\{0,1\}\big\}$, with $i\!\in\!\{1...|\mathcal{D}|\}$. Here $y_i$ denotes whether relation $r_i \in \mathcal{R}$ holds between $h_i, t_i \in \mathcal{E}$. Each entity $e\!\in\!\mathcal{E}$ is encoded as a vector $\textbf{v}_e\!\in\!\mathbb{R}^{d_\mathcal{E}}$, and each relation $r\!\in\!\mathcal{R}$ is encoded as a mapping between vectors $\textbf{W}_r\!\in\!\mathbb{R}^{d_\mathcal{R}}$, where $d_\mathcal{E}$ and $d_\mathcal{R}$ are the dimensions of vectors and mappings respectively~\cite{wang2017kge_survey,nickel2016review}. The embeddings for $\mathcal{E}$ and $\mathcal{R}$ are typically learned using a scoring function $f(h,r,t)$ that assigns higher (lower) values to positive (negative) triples~\cite{nickel2016review}. The learning objective is thus to find a set of embeddings $\Theta = \big\{\{\textbf{v}_{e}|\,e\in\mathcal{E}\},\{\textbf{W}_{r}|\,r\in\mathcal{R}\}\big\}$ that minimize the loss $\mathcal{L}$ over the training split $\mathcal{D}_{Tr}$ of the dataset $\mathcal{D}$. Loss $\mathcal{L}_\mathcal{D}$ can take many forms depending on the KGE representation used, e.g., Margin-Ranking Loss~\cite{bordes2013translating} or Negative Log-Likelihood Loss~\cite{liu2017analogical}. The learned embeddings $\Theta$ are used to infer the likelihoods of facts in the held out splits of the dataset (i.e. $\mathcal{D}_{Va}$ and $\mathcal{D}_{Tr}$), which are not present in the training split.

We use the Analogy embedding representation as in \cite{daruna2019robocse}. Analogy represents relationships as (bi)linear mappings between entities, i.e., $\textbf{v}^{\top}_{h}\textbf{W}_{r}=\textbf{v}^{\top}_{t}$~\cite{liu2017analogical}. It uses the scoring and negative log loss functions in Equations~\ref{eq:analogy_score} and~\ref{eq:analogy_loss} where $\sigma$ is a sigmoid function, $y$ is a label indicating whether the triple is corrupted, and $\mathcal{G}'$ is the corrupted knowledge graph containing only false triples. Additionally, the linear mappings (i.e. relations) are constrained to form a commuting family of normal mappings, i.e., $\textbf{W}_{r}\textbf{W}^{\top}_{r}=\textbf{W}^{\top}_{r}\textbf{W}_{r}\,\forall\,r\in\mathcal{R}$ and $\textbf{W}_{r}\textbf{W}_{r'}=\textbf{W}_{r'}\textbf{W}_{r}\,\forall\,r,r'\in\mathcal{R}$, to promote analogical structure within the embedding space.
\vspace{-0.2cm}
\begin{equation}
    \small f(h,r,t) = \langle\textbf{v}^{\top}_{h}\textbf{W}_{r},\textbf{v}_{t}\rangle
\label{eq:analogy_score}
\vspace{-0.4cm}
\end{equation}
\begin{equation}
\small \mathcal{L} = \sum\limits_{\small (h,r,t,y)\in\,{\mathcal{G},\mathcal{G'}}}
-\textrm{log}\,\sigma(y \cdot f(h,r,t))
\label{eq:analogy_loss}
\vspace{-0.2cm}
\end{equation}
In this work we extend the original knowledge graph presented in \cite{daruna2019robocse} by adding new relation types and entity types. We added 9 new relation types and a new entity type that represents action effects. Triples were extracted from VirtualHome simulations as in \cite{daruna2019robocse}, using the expanded set of relations and entities. Table~\ref{tbl:dataset} presents examples and statistics about the newly extracted knowledge graph from VirtualHome. Our experimental evaluations show that these additions to the knowledge graph significantly improve the generalization capabilities of the task generalization module against the prior knowledge graph.
\vspace{-0.1cm}

\begin{table}
    \setlength{\tabcolsep}{4pt} 
    \vspace{0.1cm}
	\caption{\centering \small Dataset extracted from VirtualHome to learn $\mathcal{G}$}
	\vspace{-0.2cm}
    \resizebox{\columnwidth}{!}{
        \begin{tabular}{l|ccccc} 
            \toprule
            Relation & $|E_{head}|^{\dagger}$ & $|E_{tail}|^{\dagger}$ & $|\mathcal{D}_{Tr}|^{\ddag}$ & $|\mathcal{D}_{Va}|/|\mathcal{D}_{Te}|^{\dagger}$ & $|\mathcal{D}|^{\dagger}$ \\ 
            \midrule
            HasEffect       & 28 & 10   & 39,263~(24) & 2 & 28 \\ 
            InverseActionOf & 10 & 10    & 29,956~(10) & 1 & 12 \\
            InverseStateOf  & 15 & 15   & 23,763~(13) & 1 & 15 \\  
            LocInRoom       & 41 & 4    & 3,972~(78) & 9 & 96 \\ 
            ObjCanBe        & 159 & 33  & 45,075~(886) & 110 & 1106 \\
            ObjInLoc        & 164 & 28  & 9,461~(409) & 51 & 511 \\ 
            ObjInRoom       & 164 & 4   & 8,276~(289) & 36 & 361 \\ 
            ObjOnLoc        & 149 & 32  & 2,346~(269) & 33 & 335 \\ 
            ObjUsedTo       & 71 & 21   & 6,224~(76) & 9 & 94 \\ 
            ObjHasState     & 153 & 15  & 28,306~(431) & 53 & 537 \\ 
            OperatesOn      & 68 & 125  & 84,286~(1124) & 140 & 1404 \\
            \midrule
            \multicolumn{6}{c}{Example entities (281 total entities)} \\
            \midrule
            \small Rooms (4) & \multicolumn{5}{c}{\small kitchen, bedroom, bathroom, livingroom} \\
            \small Locations (45) & \multicolumn{5}{c}{\small fridge, table, sink, garbage, bed, desk, cabinet, drawer} \\
            \small Objects (182) & \multicolumn{5}{c}{\small chair, towel, bleach, tomato, rug, plant, fork, laptop} \\
            \small Actions (34) & \multicolumn{5}{c}{\small wipe, open, pick up, turn off, bake, unplug, disinfect} \\
            \small States (16) & \multicolumn{5}{c}{\small dirty, clean, on, off, cooked, broken, open, plugged in} \\
            \bottomrule
        \end{tabular}
        \label{tbl:dataset}
    }
    \footnotesize{\hspace*{\fill} $^{\dagger}|\textrm{unique instances}|, ^{\ddag}|\textrm{instances}| (|\textrm{unique instances}|)$}
    \vspace{-0.7cm}
\end{table}

\subsection{Task Generalizations}
\label{sec:generalization}

We focus on generalizing object-oriented primitive actions in this work, and denote an action $a^i_d \in T_d$ associated with an object $o_d$ present at a location $l_d$, as\footnote{In this notation, any action $a^i_d(o_d, l_d)$, e.g., ``Scrub(scrubber, shelf)'', is \textit{not} the parameterization of the action itself. Instead, it indicates that the action ``scrub'' uses the object ``scrubber'', which is found on the ``shelf''.} $a^i_d(o_d, l_d)$. We assume object oriented primitive actions fail for 3 types of reasons, $o_d$ not being found in the demonstrated location $l_d$, $o_d$ not being present in the environment to perform $a^i_d$, or the environment lacking any object that can be used to perform $a^i_d$. These three failure types lead to three levels of reasoning about the task plan constituents, that can be combined and used to generalize the demonstrated task plan. We categorize these generalizations based on the specific variable that each is associated with, namely, $l_{d}$, $o_{d}$ and/or $a_{d}$. With each reasoning level, the robot generalizes up to 1, 2 or 3 types of variables:

\begin{small_ind_enumerate}
    \item \textbf{Spatial Reasoning}: This reasoning level only requires spatial generalizations to generate an executable plan.
        \begin{small_ind_s_itemize}
            \item Location ($L$): These generalizations arise when $o_{d}$ is in a different location than $l_{d}$ in the execution environment. Other likely locations for $o_{d}$ are inferred. 
        \end{small_ind_s_itemize}
    \item \textbf{Object Reasoning}: This reasoning level requires object reasoning and potentially, spatial generalizations.
        \begin{small_ind_s_itemize}
            \item Object ($O$): These generalizations arise when $o_{d}$ is not available in the execution environment, but an appropriate substitute object that can be used to perform $a_d$ is available at the location $l_{d}$. Other appropriate objects to perform $a_{d}$ are inferred. 
            \item Object-Location ($OL$): These generalizations like $O$ lack the demonstrated object $o_{d}$, but the appropriate substitute is not in the demonstrated location $l_{d}$. Other appropriate objects to perform $a_{d}$ are inferred as well as their corresponding likely locations.
        \end{small_ind_s_itemize}
    \item \textbf{Action Reasoning}: This reasoning level requires both action and object generalizations, and potentially, spatial generalizations as well. 
        \begin{small_ind_s_itemize}
            \item Action-Object ($AO$): These generalizations arise when all objects associated with $a_{d}$ are unavailable in the execution environment. An appropriate action that achieves the same effect as $a_{d}$, and a corresponding object that is a valid parameterization of the action must be inferred.
            \item Action-Object-Location ($AOL$): These generalizations are similar to $AO$, but additionally, the substitute object is not available at the demonstrated location. In this case, an appropriate action, object, and location must be inferred to generalize the task plan.
        \end{small_ind_s_itemize}
\end{small_ind_enumerate}

In this work, we make a simplifying assumption that actions with the same effects have the same preconditions. The implications of this assumption are lessened by using auxiliary queries to the knowledge graph $\mathcal{G}$ that infer which demonstrated objects are applicable to the task. Our future work will relax this assumption to learn and infer the full parameterizations of actions. In the following section, we describe the task generalization module's implementation of the three generalizations above that integrates the knowledge representation in Section~\ref{sec:KGE} with the task plan.

\subsection{Task Generalization Module}
\label{sec:module}

\setlength{\textfloatsep}{0.1cm}
\setlength{\floatsep}{0.1cm}
\begin{algorithm}[t]
    \SetKwInOut{Input}{input}\SetKwInOut{Output}{output}
	\DontPrintSemicolon
    \SetKwFunction{FMain}{\small generalize\_task}
    \SetKwProg{Fn}{\small Function}{:}{}
    \Fn{\FMain{\small $T$}}{
        \small $l_{f}$, $o_{f}$, $a_{f}$ = extract\_failure(T)
    
        \uIf{\small $l_{subs}$ {\rm {\bf is}} $\emptyset$} {
            \small $l_{subs}$ = infer $l_{subs}$ for $o_{f}$ from $\mathcal{G}$
        }
        \small $l$ = {\rm {\bf next}}$(l_{subs})$
        
        \uIf(\tcp*[h]{\small Spatial Gen.}){\small $l \neq {\rm None}$} {
            \uIf{\small $\exists l$}{
                \small $T_x$ = {\rm replace}$(l_f, l, T)$
                
                \small return $T_x$
            } \uElse {
                \small {\rm {\bf goto}} line 5
            }
        }
        
        \uIf{\small $o_{subs}$ {\rm {\bf is}} $\emptyset$} {
                \small $o_{subs}$ = infer $o_{subs}$ used to $a_{f}$ from $\mathcal{G}$
        }
        
        \small $o$ = {\rm {\bf next}}$(o_{subs})$
        
        \small $l_{subs} = \emptyset$
        
        \uIf(\tcp*[h]{\small Object Gen.}){\small $o \neq {\rm None}$}{
            \small $T$ = {\rm replace}$(o_f, o, T)$
            
            \small $o_f$ = $o$
            
            \small {\rm {\bf goto}} line 3
        }
        
        \uIf{\small $a_{subs}$ {\rm {\bf is}} $\emptyset$} {
            \small $a_{subs}$ = infer $a_{subs}$ with {\rm effect($a_f$)} from $\mathcal{G}$
        }
        \small $a$ = {\rm {\bf next}}$(a_{subs})$
        
        \small $o_{subs} = \emptyset$
        
        \uIf(\tcp*[h]{\small Action Gen.}){\small $a \neq {\rm None}$}{
            \small $T$ = {\rm replace}$(a_f, a, T)$
            
            \small $a_f$ = $a$
            
            \small {\rm {\bf goto}} line 12
        }
        
        \small return {\rm Task Failure}
    }
	\caption{\small Task Generalization Algorithm}
	\label{alg:1}
\end{algorithm}

Algorithm~\ref{alg:1} shows our approach to one-shot task execution. Algorithm~\ref{alg:1} implements the task generalizations in Section~\ref{sec:generalization} using the computational framework in Section~\ref{sec:KGE}. The pseudocode for Algorithm~\ref{alg:1} is written as a callable Python object with three queues as class members that maintain state across function calls. Algorithm~\ref{alg:1} incrementally updates the failed task plan passed as the input parameter $T$, which is initially the demonstrated task plan $T_{d}$. The incremental task plan updates and executions continue until an executable task plan $T_{x}$ is found or the algorithm cannot find an executable task generalization for the environment (i.e. returns Task Failure). Each incremental update to the task plan $T$ is achieved by replacing all mentions of the variable that is being generalized in the task plan $T$, denoted by the {\it replace} function. Algorithm~\ref{alg:1} was implemented to prioritize generalizations involving fewer variables (i.e. generalizing one type of variable is preferable to two). This approach leads to simpler generalizations that more closely match the demonstrated task plan.

Algorithm~\ref{alg:1} uses the three levels of generalization discussed in Section~\ref{sec:generalization} depending on the variable being reasoned about (location, object, or action). Each level of reasoning uses different sets of queries to infer generalizations from the knowledge representation. The KGE infers location generalizations by completing the triples $(o_f , {\rm ObjAtLoc}, \rule{3mm}{0.15mm})$ and $(o_f , {\rm ObjInRoom},\rule{3mm}{0.15mm})$. The inference results of these incomplete triples are combined to produce a prioritized queue of each likely location instance in the environment for $o_{f}$. Similarly, the KGE infers object generalizations by completing the triples $(o_f , {\rm ObjUsedTo}, \rule{3mm}{0.15mm})$ and $(o_f , {\rm OperatesOn}, \rule{3mm}{0.15mm})$. The inference results of these incomplete triples are combined to produce a prioritized queue of each viable object class that can perform the desired action on the target object. Lastly, when inferring action generalizations, the KGE completes the triple $(\rule{3mm}{0.15mm}, {\rm ActionHasEffect}, {\rm effect}(a_{f}))$. The inference results of this incomplete triple produce a prioritized queue of each viable action class that achieves the desired effect. The queue for each reasoning level has a fixed size, which is tuned and discussed further in Section~\ref{sec:exp2}.

\section{Experimental Evaluation}
\label{sec:exp_pro}

Our experimental evaluations model the task domain of cleaning since it is a commonly desired task \cite{cakmak2013towards} and involves a variety of objects and actions. We perform evaluations both in simulation using VirtualHome~\cite{puig2018virtualhome}, and on a physical robot (Fetch~\cite{wise2016fetch}). Our {\it simulation} experiments included 7 cleaning actions (``wipe'', ``dust'', ``sweep'', ``wash'', ``rinse'', ``disinfect'', and ``scrub''), 28 cleaning objects, and 45 possible object locations leading to 8,820 possible generalizations for a primitive action $a^i_d(o_d, l_d)$. Our {\it robot} experiments included 4 cleaning actions (``wipe'', ``dust'', ``sweep'', and ``scrub''), 5 cleaning objects, and 12 possible object locations leading to 240 possible generalizations for a primitive action $a^i_d(o_d, l_d)$. The large number of possible task generalizations in both cases highlights the potential of incorporating learned domain knowledge into the task demonstration, as well as the challenge when reasoning about the most likely generalizations given the scale and complexity of probabilistic distributions within the domain.

We performed 3 evaluations; first, to broadly compare existing knowledge representations against ours, second to further analyze components of Algorithm~\ref{alg:1} in ablation studies, and last to validate our approach on a physical robot. We tested the statistical significance of results using repeated-measures ANOVA and a post-hoc Tukey’s test.

\subsection{Comparison of Knowledge Representations}
\label{sec:exp1}

In this experiment we focus on comparing our knowledge representation to baselines representative of prior work in the context of robust task execution. The baselines were:
\begin{small_ind_enumerate}
    \item {\bf Single Demo (SD)} serves as a reference for how often the original demonstration is randomly spawned.
    \item {\bf Plan Network (PN)} answers queries by building a repository of valid location, object, and action substitutes observed across consecutive executions, similar to \cite{orkin2007restaurant}. Repository is built from demonstrations obtained from an oracle (i.e. solutions of test execution environments).
    \item {\bf Word Embedding (WE)} answers queries using cosine similarity over ConceptNet Numberbatch v19.08 embeddings \cite{speer2016conceptnet}. Word embeddings have been used in robotics works as semantic representations \cite{fulda2017harvesting,thomason2018guiding,scalise2019improving}.
    \item {\bf Markov Logic Network (MLN)} answers queries by learning statistical correlations relating objects and object attributes as in \cite{nyga2018grounding,zhu2014reasoning}  trained on the dataset in Table~\ref{tbl:dataset}.
    \item {\bf Original RoboCSE (RCSE)} answers queries using the KGE framework in~\cite{daruna2019robocse} that lacks the entity and relation types added in this work. Trained on the dataset in Table~\ref{tbl:dataset}.
    \item {\bf Training Set Memorization (TM)} answers queries using only training triples in Table~\ref{tbl:dataset} without a KGE (i.e. lacks inference of valid/test triples).
\end{small_ind_enumerate}

\begin{table}
    \setlength{\tabcolsep}{4pt} 
    \vspace{0.1cm}
	\caption{\centering \small Metrics for 12,000 Environment Executions}
	\vspace{-0.2cm}
    \resizebox{\columnwidth}{!}{
        \begin{tabular}{l|cccccc} 
        \toprule
        \midrule
        Metric & SD & PN & WE & RCSE & TM & Ours \\
        \midrule
        Success Rate (\%) & 1.2$\pm$1.3 & 54.6$\pm$2.6 & 9.8$\pm$7.8 & 19.9$\pm$6.1 & 68.8$\pm$3.3 & {\bf 73.7$\pm$2.6} \\ 
        Num. Attempts    & 1$\pm$0 & 17$\pm$14 & 30$\pm$33 & 62$\pm$34 & 32$\pm$14 & 67$\pm$28 \\
        \bottomrule
        \end{tabular}
        \label{tbl:exp1}
    }
\end{table}

\begin{figure}[t]
	\centering
	\includegraphics[width=0.9\linewidth]{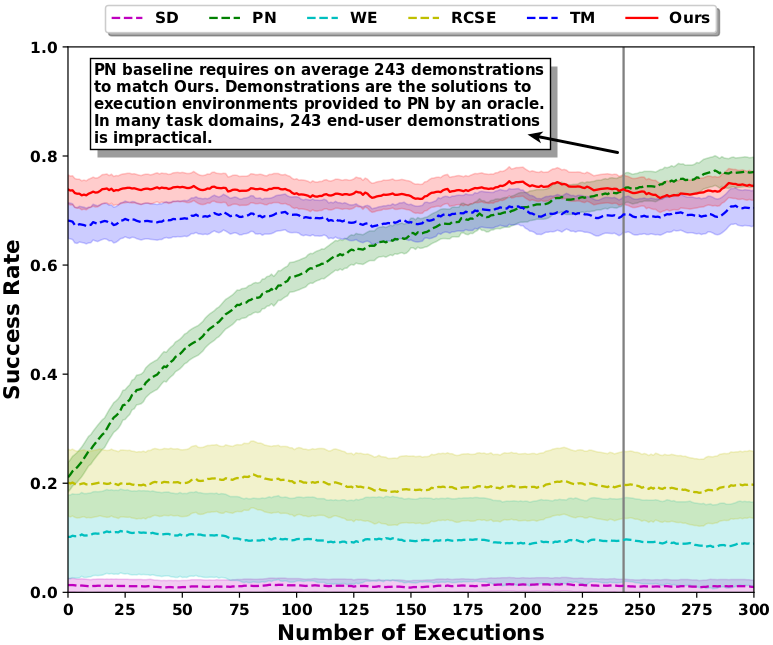}
	\vspace{-0.2cm}
	\captionsetup{width=\linewidth}
	\caption{\small Moving average success rate of 60 executions for all tested task generalization approaches.}
	\label{fig:exp1_line}
\end{figure}

Each simulation experiment in VirtualHome begins by spawning an agent in an initial environment $E_d$. A hierarchical task network provides the agent with a task plan $T_d$ in the initial environment $E_d$ that serves as the single task demonstration. The demonstrated task plan $T_d$ defines an initial cleaning task by performing a cleaning action $a_d$ using a demonstration object $o_d$ found in a demonstrated location $l_d$. Then, the agent must execute the task plan $T_d$ in an execution environment $E_x$, which is a perturbation of the demonstration environment $E_d$. If a primitive action fails when executing $T_d$, the agent iteratively invokes the task generalization module until it finds an executable task plan $T_x$ or fails. We track the {\it number of attempts} the agent makes at completing the task plan as well as whether it {\it succeeds}. The agent must attempt to generalize the demonstrated task plan to 300 other randomly generated execution environments. In this experiment, execution environments are generated by perturbing the object used to demonstrate the task; changing the object's location, type, or both. The perturbations are made in accordance with the action-object-location distributions present in VirtualHome, ensuring that objects are not placed at implausible locations (e.g., broom inside the toilet) and that the intended generalization is not unreasonable (e.g. cleaning a table with a washing-machine). Each execution environment has exactly one valid solution. We repeat the experiment 40 times with different initial demonstrations to avoid over-fitting reported results to an initial demonstration that is an outlier, totalling 12,000 sampled execution environments. Note that the domain knowledge representation was the only variable across the experiments, while other components of the system architecture in Figure~\ref{fig:Pipeline}, initial demonstrations, and perturbations were controlled. For all methods, we tuned the sizes of prioritized queues in Algorithm~\ref{alg:1} to 12 for locations, 8 for objects, and 4 for actions using an analysis described in Section~\ref{sec:exp2}.

Table~\ref{tbl:exp1} and Figure~\ref{fig:exp1_line} summarize the major results of this experiment. Table~\ref{tbl:exp1} shows the success rate and number of attempts for each approach averaged across the 40 experiment trials. As shown in Table~\ref{tbl:exp1}, Ours provides improvements in success rate compared against all the baselines with {\it strong statistical significance} (i.e. $p<0.001)$. Without the capacity to directly model relations, WE produces less relevant generalizations using word similarities to the initial demonstration. RCSE's disadvantage compared with Ours stems from the lack of modeling action effects as well as other relationships that can help to rank more relevant generalizations higher. Crucially, Ours also outperforms TM, signaling that the embedding is inferring generalizations not possible with the training set alone. Lastly, we can consider the PN baseline; however, analyzing Table~\ref{tbl:exp1} hides the transient behavior of PN as it accumulates demonstrations. We provide the moving average over a window of 60 executions in Figure~\ref{fig:exp1_line} to show how PN eventually approaches the performance of Ours, but only after PN accumulates on average 243 additional demonstrations from the oracle. PN highlights a limitation of Ours, the lack of adaptation to execution outcomes leading to a flat success rate. However, despite PN having the additional 243 demonstrations, Ours outperforms PN on average with a single demonstration because Ours learns priors over the semantic domain knowledge in VirtualHome.

Unlike our KGE, the MLN baseline implemented with pracmln~\cite{nyga2013pracmln} is not included in the above results because the scale of the learned knowledge graph (see Table~\ref{tbl:dataset}) caused intractability issues. Despite simplifying the problem from 868,571 ground variables\footnote{$\mathcal{O}(|L|*|C|^{N})$ where $|L|$ is the number of predicates (relations - 11), $|L|$ is the number of constants (entities - 281), and $N$ is the maximum arity of a predicate in symbol space (2).} to 67,656, MC-SAT required 10 minutes to perform full-posterior inference for a single ground atom. These inferences occur many times during a single sample of the 12,000 sampled execution environments making the MLN baseline impractical for our use case.

\subsection{Ablation Studies of Algorithm~\ref{alg:1}}
\label{sec:exp2}

We now provide multiple ablation studies of Algorithm~\ref{alg:1}.

{\bf Characterizing Algorithm~\ref{alg:1} reasoning levels:} We implemented 5 ablations that allowed combinations of generalizations at the spatial, object, or action reasoning levels. These 5 ablations implement at least one of the 5 types of generalizations presented in Section~\ref{sec:generalization} by toggling portions of Algorithm~\ref{alg:1} (i.e. O, L, OL, AO, AOL). In addition, we perturbed environments in a controlled manner to observe which components of our algorithm were suited to which environment perturbation types. The environment perturbation types corresponded to one of the 5 generalization cases in Section~\ref{sec:generalization} (e.g. for L, only perturb location of $o_{d}$) or random perturbations as in the prior evaluation of Section~\ref{sec:exp1}. We reduced the number of initial demonstrations to 10 and execution environments to 50 due to the large number of cases for each ablation and test environment combination. Our results, shown in Table \ref{tbl:ablation}, highlight the importance of combined reasoning at multiple levels to generalize the demonstrated task plan to larger number of execution environments.

{\bf Characterizing Algorithm~\ref{alg:1} queue sizes:} We defined ablations by varying the queue sizes for each reasoning level in proportion to the total number of valid generalizations (i.e. $|l_{subs}|\!=\!\{6,12,24\}$, $|o_{subs}|\!=\!\{4,8,16\}$, $|a_{subs}|\!=\!\{2,4,8\}$), leading to a total of 27 different ablations. Random environment perturbations were used for testing, as in the prior evaluation of Section~\ref{sec:exp1}. Due to the large number of ablations, we again reduced the number of initial demonstrations to 10 and execution environments to 50. We selected $\{|l_{subs}|,|o_{subs}|,|a_{subs}|\}\!=\!\{12,8,4\}$ from the results of this grid search to have moderate success rates while keeping the number of generalization attempts below 1\% of the total possible (8,820). Overall, the different ablations indicated a trade-off for priority queue size; larger queue sizes lead to higher success rates but also higher numbers of generalization attempts.

\subsection{Validation on a Robot Platform}
\label{sec:exp3}

Our last experiment was to validate our approach on a physical robot platform, Fetch \cite{wise2016fetch}, using a task execution stack that facilitates execution of primitive actions \cite{banerjee2020taking}. The task execution stack was used to make the Fetch execute task plans, receive environment percepts indicating failed primitive actions, and accept generalizations from the task generalization module. Some errors outside the scope of our work are handled by the task execution stack (e.g. base path-planning obstacles, precarious grasps); however, any errors not handled by the task execution stack or our approach are considered failures in our test cases, as all executions are autonomous without human intervention\footnote{Example executions: \url{https://youtu.be/epRjleYDTCw}}.

\begin{figure}[t]
	\centering
	\includegraphics[width=0.3\textwidth]{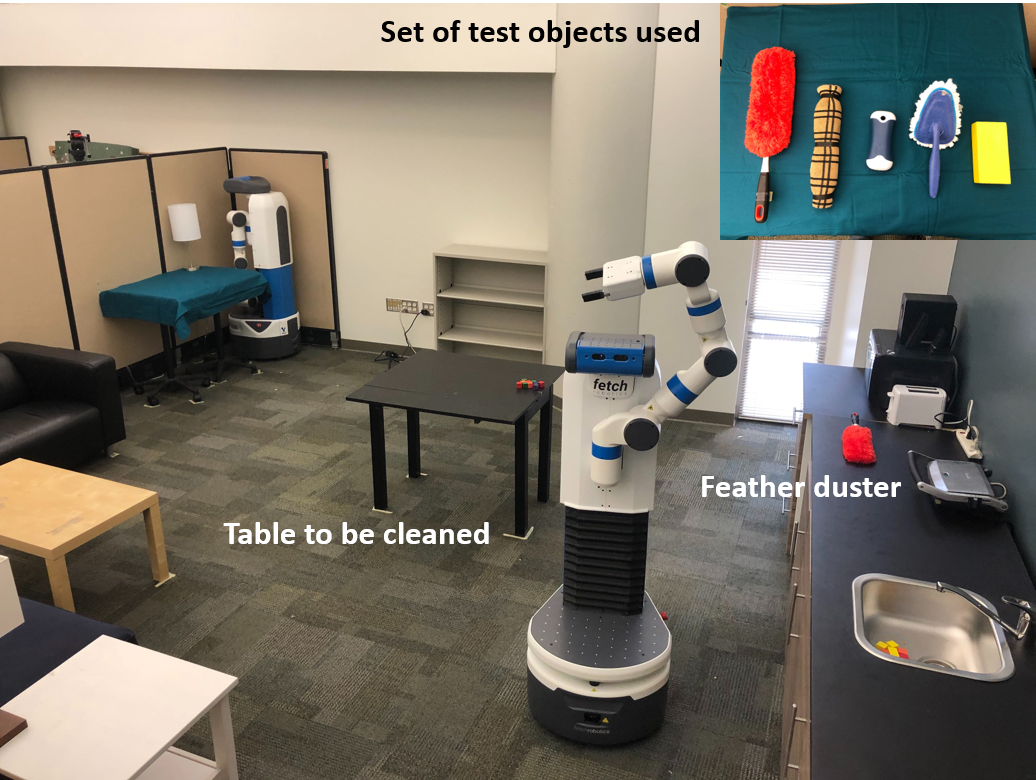}
	\captionsetup{width=\linewidth}
	\caption{\small Sample test environment set up similar to a household setting. Test objects shown in top right, from left to right: feather duster, towel rolled, scrubber, duster, and washing sponge.}
	\label{fig:robot_expt}
\end{figure}

In our robot experiments, we used five different objects, namely ``towel rolled'', ``washing sponge'', ``scrubber'', ``feather duster'' and ``duster'' (Shown in Figure \ref{fig:robot_expt}), with four different actions ``wipe'', ``scrub'', ``dust'' and ``sweep''. Possible locations for objects include, kitchen counters, coffee tables, sofa, desk, sink, drawers, shelf and kitchen table. The testing environment that emulates a small studio apartment is shown in Figure \ref{fig:robot_expt}. As shown in the figure, the robot has to find an appropriate object in the environment to clean the designated location. We evaluated our approach on the robot by generating 10 random perturbations of one cleaning task demonstrated for each object, as described in Section~\ref{sec:exp1}.

\begin{table}
    \setlength{\tabcolsep}{4pt} 
    \vspace{0.1cm}
	\caption{\centering \small Ablation Success Rates in 6 $E_{x}$ Types}
	\vspace{-0.2cm}
    \resizebox{\columnwidth}{!}{
        \begin{tabular}{l|ccccc} 
        \toprule
        $E_{x}$ Type & Abl. L & Abl. O & Abl. OL & Abl. AO & Abl. AOL \\
        \midrule
        L Gen. Only     & 86.4\% & 0.0\% & 86.4\% & 0.0\% & 86.4\% \\ 
        O Gen. Only     & 0.0\% & 71.6\% & 71.6\% & 82.2\% & 82.2\% \\ 
        OL Gen. Only    & 0.0\% & 0.0\% & 68.2\% & 0.0\% & 74.6\% \\ 
        AO Gen. Only    & 0.0\% & 0.0\% & 0.0\% & 80.2\% & 80.2\% \\ 
        AOL Gen. Only   & 0.0\% & 0.0\% & 0.0\% & 0.0\% & 67.4\% \\ 
        Random Gen.     & 6.2\% & 4.6\% & 24.4\% & 14.8\% & 71.8\% \\ 
        \bottomrule
        \end{tabular}
        \label{tbl:ablation}
    }
\end{table}

In our robot evaluations, the robot successfully generalized demonstrated tasks to 38 of 50 total execution environments (76\% success rate). We note that none of the failures were due to our approach. The failure cases were due to manipulation (2), object detection (7), plane-segmentation (1), and grasping (2) robot errors. It is worth noting the KGE inferred three key triples not within the training set used to learn the embedding (i.e. (featherduster, ObjOnLoc, shelf), (duster, ObjUsedTo, dust), (towelrolled, ObjUsedTo, wipe)) to generate 10 of the 38 successful generalizations. This result highlights the benefit of using a knowledge representation that can infer unseen facts given existing facts.

\section{Conclusion}
\label{sec:conclusion}

Robust one-shot task execution continues to be a challenging problem for learning from demonstration. We introduced the task generalization module for generalization of task plans to new execution environments by integrating task plans with a KGE learned from observations in a simulator. We compared our approach to representative baselines in the context of one-shot task execution in simulation. Our experiments demonstrated that our knowledge representation infers more relevant generalizations on average, leading to higher success rates than the baselines. Lastly, we validated our work on a physical platform, showing generalization of the demonstrated tasks to 38/50 execution environments, including technical failures. Our future work will explore how the domain knowledge can adapt to outcomes observed during execution, instead of solely relying on the prior distributions as is common in existing work.


\bibliographystyle{./IEEEtran}
\balance
\bibliography{references}  

\end{document}